\documentclass[letterpaper, 10 pt, conference]{ieeeconf}  

\IEEEoverridecommandlockouts                              

\usepackage[utf8]{inputenc}
\usepackage{graphics} 
\usepackage{epsfig} 
\usepackage{mathptmx} 
\usepackage{times} 
\usepackage{amsmath} 
\usepackage{amssymb}  
\usepackage{textcomp}
\usepackage{siunitx}
\usepackage{xspace}
\usepackage[tracking=true]{microtype} 

\usepackage{caption}
\usepackage[dvipsnames]{xcolor}
\usepackage[url=false,isbn=false,doi=true,style=ieee,backend=biber,maxnames=5,date=year,seconds=true,autocite=inline]{biblatex}
\let\labelindent\relax
\usepackage[inline]{enumitem}
\usepackage{booktabs}
\usepackage[overload]{textcase}
\usepackage[caption=false]{subfig}

\usepackage{ifthen}

\makeatletter
\newcounter{IEEE@bibentries}
\renewcommand\IEEEtriggeratref[1]{%
  \renewbibmacro{finentry}{%
    \stepcounter{IEEE@bibentries}%
    \ifthenelse{\equal{\value{IEEE@bibentries}}{#1}}
    {\finentry\@IEEEtriggercmd}
    {\finentry}%
  }%
}
\makeatother
\AtBeginBibliography{\footnotesize}
\newcommand{\term}[1]{\textit{#1}}
\newcommand{\principle}[1]{\textit{#1}}
\newcommand{\abbr}[1]{\textls[50]{\MakeTextUppercase{#1}}}
\newcommand{\hri}{\abbr{hri}\xspace}

\newcommand{\siso}{slow in, slow out\xspace}

\newcommand{\SiSo}{Slow in, Slow out\xspace}

\title{\LARGE \bf
Differences of Human Perceptions of a Robot Moving using Linear or \SiSo Velocity Profiles When Performing a Cleaning Task
}

\author{Trenton Schulz$^{1}$, Patrick Holthaus$^{2}$, Farshid Amirabdollahian$^{2}$, Kheng Lee Koay$^{2}$,\\
  Jim Torresen$^{1}$, and Jo Herstad$^{1}$
\thanks{*Partly funded by the Research Council of Norway as part of the Multimodal Elderly Care Systems (\abbr{mecs}) project, under grant agreement~247697.}
\thanks{$^{1}$Department of Informatics,
        University of Oslo, Postbox 1080 Blindern, 0316 Oslo
        {\tt\small [trentonw|jimtoer|johe]@ifi.uio.no}}%
\thanks{$^{2}$School of Computer Science, University of Hertfordshire, Hatfield, England
        {\tt\small [p.holthaus|f.amirabdollahian2 |k.l.koay]@herts.ac.uk}}%
}

\begin{document}

\maketitle
\thispagestyle{empty}
\pagestyle{empty}


\begin{abstract}
  We investigated how a robot moving with different velocity profiles
  affects a person's perception of it when working together on a
  task. The two profiles are the common linear profile and a profile
  based on the animation principles of \siso.  The investigation was accomplished
  by running an experiment in a home context where people and the
  robot cooperated on a clean-up task. We used the Godspeed series of
  questionnaires to gather people's perception of the robot. Average
  scores for each series appear not to be different enough to reject
  the null hypotheses, but looking at the component items provides
  paths to future areas of research. We also discuss the scenario for
  the experiment and how it may be used for future research into using
  animation techniques for moving robots and improving the legibility
  of a robot's locomotion.
\end{abstract}

\section{Introduction}\label{sec:introduction}

One way older, retired people can live independently
at home longer is to have robots in the home to help them with tasks.
People interacting with a robot in a home environment  will need
to perceive the robot as trustworthy, understandable, and something
they may want to be around. Can the way a robot speeds up and slows down
affect when moving a person's feeling of safety and perception of the robot?

\begin{figure}[htb]
\centering
\includegraphics[trim={0cm 5.9cm 0cm 6.0cm}, clip, width=0.49\textwidth]{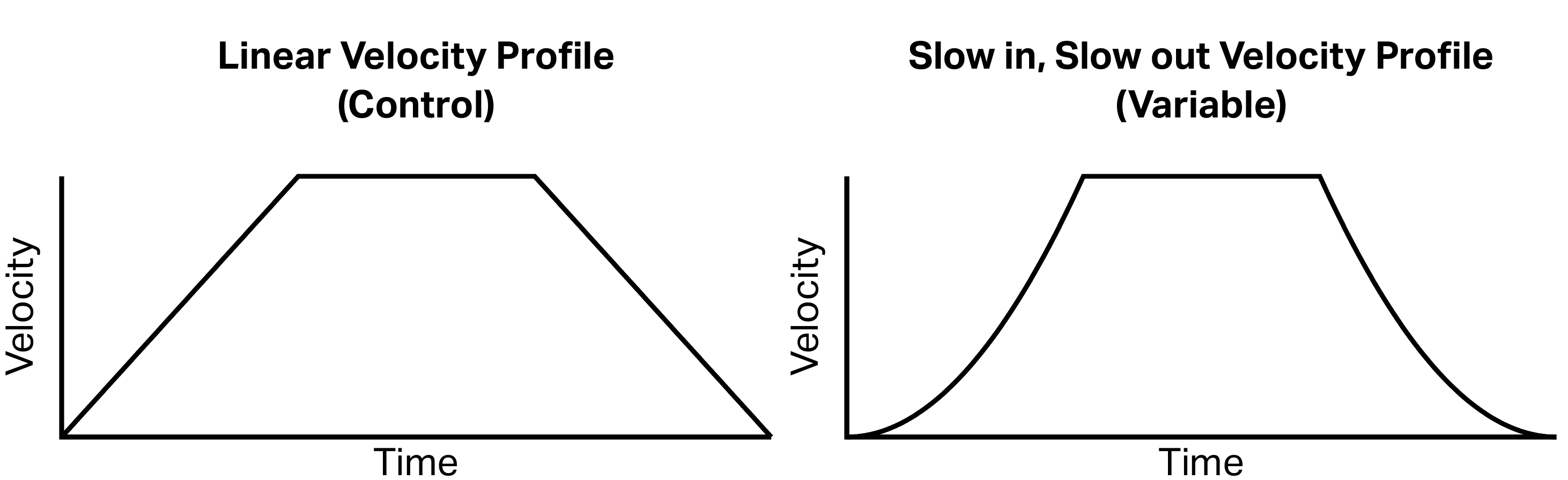}
\caption{The experiment tests how people's perception are affected by a robot that moves with a linear velocity profile (\emph{left}) versus a velocity profile that is based on the animation principle of \siso (\emph{right}).}\label{fig:velocity-profiles}
\end{figure}

\term{Velocity profiles} (or velocity curves) describe the
acceleration of a robot. Most robots use a velocity profile that
produces linear acceleration (Fig.~\ref{fig:velocity-profiles},
\emph{left}), which results in a mechanical, robot-like movement, but
there are other velocity profiles available. One of the principles of
film animation is the idea of \term{\siso}
\parencite{ThomasIllusionLifeDisney1995}. In \siso, the movement
starts slowly, but then speeds up to its top speed and slowly stops as
it reaches its destination (Fig.~\ref{fig:velocity-profiles},
\emph{right}). The goal behind \siso in film animation is to make a
character's movement seem more realistic (that is, like something that
is alive and people are familiar with). Applying a \siso velocity
profile to a robot may make the robot's movement more legible and
easier to follow resulting in people having a different perception of
the robot than if it simply used the linear velocity profile.

In earlier work \parencite{SchulzClassifyingHumanRobot2018}, we
implemented an algorithm for changing a ground-based robot's velocity
profile to create \siso movement, but had yet to test how this
affects people's perception of a robot, especially when people
interact with a robot to complete a task. We wanted to investigate this
by running an experiment in a home environment. We previously documented the initial set up for the experiment and the hypotheses we wished to test \parencite{SchulzHumansPerceptionRobot2019}. Our hypotheses were \parencite[][p.~594]{SchulzHumansPerceptionRobot2019}:
\begin{enumerate*}[label=(\textit{H\arabic*})]
\item a \siso velocity profile will positively affect the robot’s perceived safety versus a linear velocity profile.
\item a \siso velocity profile will positively affect the robot’s perceived intelligence versus a linear velocity profile.
\item a \siso velocity profile will positively affect the robot’s perceived animacy versus linear a velocity profile.
\item a \siso velocity profile will positively affect the robot’s perceived anthropomorphism versus a linear velocity profile.
\item a \siso velocity profile will positively affect the robot’s perceived likability versus linear a velocity profile.
\end{enumerate*}
Here, we document the final experimental method (Section~\ref{sec:method}), present
 results (Section~\ref{sec:results}), and  discussion
(Section~\ref{sec:discussion}). But let us first examine previous human-robot interaction (\hri)
research on making robots movement legible and predictable.

\section{Related Work}\label{sec:related-work}

\Citeauthor{DraganLegibilityPredictabilityRobot2013} define
\term{predictable} movement as matching an intent of the moving
object; \term{legible} movement is where the movement could only lead
to the intent of the moving object as observed by a third party
\parencite[p.~301]{DraganLegibilityPredictabilityRobot2013}. One study
examined how a robot should approach a person to deliver a drink
\parencite{KoayExploratoryStudyRobot2007}. People preferred
different approaches and distances for the robot—sometimes
contradicting preferences from an earlier study
\parencite{WaltersRoboticEtiquetteResults2007}, but all the
participants choose movements emphasizing legibility and a lower
risk of harm to the person \parencite{KoayExploratoryStudyRobot2007}.
This is an example of predictability and legibility for locomotion.
Others have attempted to create a typology of meaningful signals and
cues for robots based on different modalities to make it easier for
people to understand information from a robot they are interacting
with \parencite{Hegeltypologymeaningfulsignals2011}.

Since a robot's arms can be heavy, powerful, and capable of harm, they
are often a focus for legibility studies (legibility for configuration).
One study found that people familiar with a robot arm's motion makes
it easier for them to predict its movement, but it was dependent on
how natural the arm's movement was
\parencite{DraganFamiliarizationRobotMotion2014}.  Another study
examined how predictable and legible movement are correlated and
formalized the difference between them into a model with the legible
motion trying to match natural human grasping motion
\parencite{DraganLegibilityPredictabilityRobot2013}. This model was
applied to a human-robot collaboration study where the legible
motion resulted in more fluid collaboration than the predictable
motion \parencite{DraganEffectsRobotMotion2015}.

Animation principles and techniques have been used make robots'
movement easier to understand to people interacting with the robot
\parencite{SchulzAnimationTechniquesHumanRobot2019}. For example, a
video study used the principles of \principle{Anticipation} and
\principle{Follow Through and Overlapping Action} to make it easier
for viewers to understand what a non-humanoid robot was intending to
do \parencite{TakayamaExpressingThoughtImproving2011}. Another study
used the \principle{timing} principle to move a robot's arm, which
helped people attribute meaning to the motion
\parencite{ZhouExpressiveRobotMotion2017}.

The \principle{\siso} principle (also known as \term{easing}) has
also been used to express intent. This principle was used to help
signal to collaborating musicians where a marimba-playing robot was
going to strike
\parencite{HoffmanInteractiveimprovisationrobotic2011}.  Another study
used three principles of animation—\principle{anticipation},
\principle{arcs}, and \principle{\siso}—to move an assistive free flyer (a type of drone) while
performing a task. Participants who watched the drone felt that using
these principles made it easier to understand the drone's intent, and
the participants also felt that the drone moved naturally and felt
safer around the drone
\parencite{SzafirCommunicationIntentAssistive2014}.  However, newer
research indicates that the motion of drones does not directly
translate to how terrestrial robots should move
\parencite{WojciechowskaCollocatedHumanDroneInteraction2019}.

Another study used a new experimental technique where participants
viewed frames of a PR2 robot that varied its movement and orientation
to see how well participants could predict movement
\parencite{PapenmeierHumanUnderstandingRobot2018}. Participants could
spend as much time viewing a frame as they wanted before proceeding to
the next. The implication being that the longer the participant looked
at the frame, the more time was spent trying to comprehend the robot's
movement, and therefore, the less natural and less predictable the
movement. The robot moved with four different velocity profiles:
linear increasing, linear decreasing, constant, and sinusoidal
(vaguely similar to \siso). Participants spent the most time viewing the
frames of a robot with the decreasing linear velocity profile, and on the
frames of where the robot decreased speed in the sinusoidal condition,
but not when it increased speed. So, the decreasing speed of the robot
affected the robot's predictability.

Our work also investigates people's perception of a moving robot and how
changing the velocity profile alters that perception. We wanted to
investigate this using a real robot in a
home context instead of a simulation or a lab setup.  We also wanted
to focus on a single animation principle to see how much power that
principle had on its own to affect people's perception of the robot.

\section{Method}\label{sec:method}

We ran a within-subjects experiment where the variable is how the
robot moves: using a linear velocity profile or a \siso velocity
profile. The aim of the experiment was to see if the velocity profiles
change people's perception of the robot as they work on a task. We did
\emph{not} want to simply ask people's opinions of the two velocity
profiles as people rarely only watch robots move. Therefore, we
created a scenario in a home environment that put the focus on the
robot interaction, and the robot's movement played a supporting role.
We first describe the procedure of the experiment and then provide
implementation details about the setting and robot.

\subsection{Experimental Procedure}

The procedure was approved by the University of Hertfordshire
Health, Science, Engineering and Technology Ethics Committee (Protocol
Number \abbr{COM}/\abbr{SF}/\abbr{UH}/03491). The procedure was as follows. First, a
consenting participant entered the home and filled out demographic
information of age, gender, and if the participant had any experience
with robots. Next, we explained the safety information about the robot
for the participant. Then, the scenario was explained: the participant
was visiting a friend’s house to help in cleaning up the home (one of
the facilitators was introduced as the friend). The robot was
also helping with the cleaning. Since we did not want to draw
attention to robot's motion, we explained we were interested in how
the robot handles the hand over of objects from the participant. Then,
the participant was given the instructions for what would happen next.

\begin{figure}[htb]
\centering
\includegraphics[trim={5cm 9.5cm 5cm 0}, clip, width=0.49\textwidth]{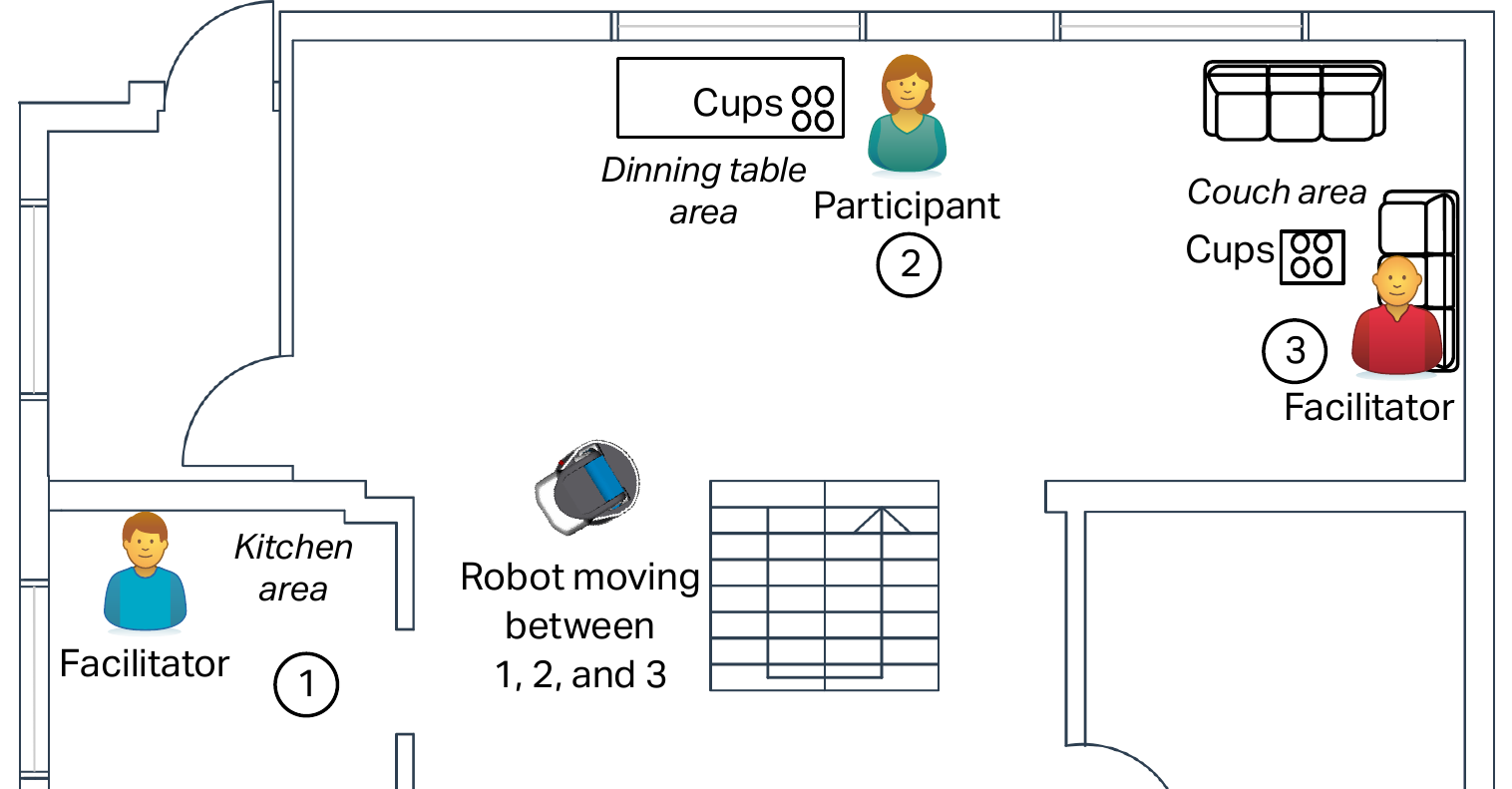}
\caption{Floor plan and position of people for the
  experiment. The robot would move between the numbered positions,
  starting at Position~1,
  using a linear or \siso velocity
  profile. }\label{fig:floor-plan}
\end{figure}

The facilitators and participant would then take their positions. This
is depicted in Fig.~\ref{fig:floor-plan}. One facilitator would stand
in the kitchen (near Position~1 in Fig.~\ref{fig:floor-plan}); the
participant and the facilitator helping in cleaning the house would
stand near the dining table and sit on the couch respectively (near
Positions~2 and 3 respectively in Fig.~\ref{fig:floor-plan}). The
dining table and the coffee table by the couch both had four cups on
them that would need to be cleaned up.

The robot started in the kitchen (at Position~1 in
Fig.~\ref{fig:floor-plan}). The procedure was as follows:
\begin{enumerate*}[label=(\textit{\alph*})]
  \item The robot moved from Position~1 to Position~2.
  \item The participant took one of the cups from the
    dining table and put it in the robot's basket.
  \item The robot moved from Position~2 to Position~3.
  \item The facilitator on the couch took a cup from the coffee table and put it in the robot's
    basket (Fig.~\ref{fig:fetch-robot}).
  \item The robot moved to Position~1.
  \item The facilitator in the kitchen removed the cups and put a copy of the
questionnaire in the basket.
  \item The robot moved to Position~2.
  \item the participant took the questionnaire from the robot and filled it out.
  \item Once the questionnaire was complete, the
    participant put the questionnaire back in the robot's basket.
  \item The robot moved to Position~1.
  \item Finally, the facilitator in the kitchen removed the questionnaire and prepared the robot for the next iteration.
\end{enumerate*}

To measure people's perceptions of the robot,
we chose the Godspeed Questionnaire \parencite{BartneckMeasurementInstrumentsAnthropomorphism2009},
which offers series  of semantic scales for measuring a robot's perceived animacy,
anthropomorphism, likeability, perceived intelligence, and perceived safety.
In addition to these scales, we included an additional item about how well
the person could predict where the robot would go, and an open question
about how well the person thought the robot performed the task.

This procedure was repeated three times. This resulted in a total of
four iterations: two times the movement was with a linear velocity
profile, and two times the movement was with a \siso velocity
profile. The profiles were counterbalanced to avoid ordering
effects. The counterbalancing was achieved by taking the six possible
combinations of two linear and two \siso velocity profiles, and
randomly selecting an ordering for each participant.

Participants were asked to stand if able while giving the cup to the
robot and receiving the questionnaire. They could sit while filling
out the questionnaire. The primary reason was to allow a better view
of Fetch and keep the base for participants' perceptions similar since
a standing participant is taller than the robot, which might not be
true with a sitting participant. A lesser, secondary reason was to
make people feel safer as the robot approached as we reasoned that
participants may feel easier to move away from a robot when they are
already standing versus having to get up from a chair.

After the final iteration, participants went through an ending
procedure where they filled out a questionnaire with open-ended
questions concerning the overall interactions. We also informed
participants that we were actually interested in the robot's movement
and not the handover. Participants could ask other questions about the
experiment and details about the robot and the house. Finally, we
thanked participants for their time and, as noted in the informed
consent form, gave them a £10 gift card for Amazon as compensation for
time and traveling to Robot House.

\subsection{Experiment Settings, Equipment, and Software}

\begin{figure}[htb]
\centering
\includegraphics[width=0.33\textwidth]{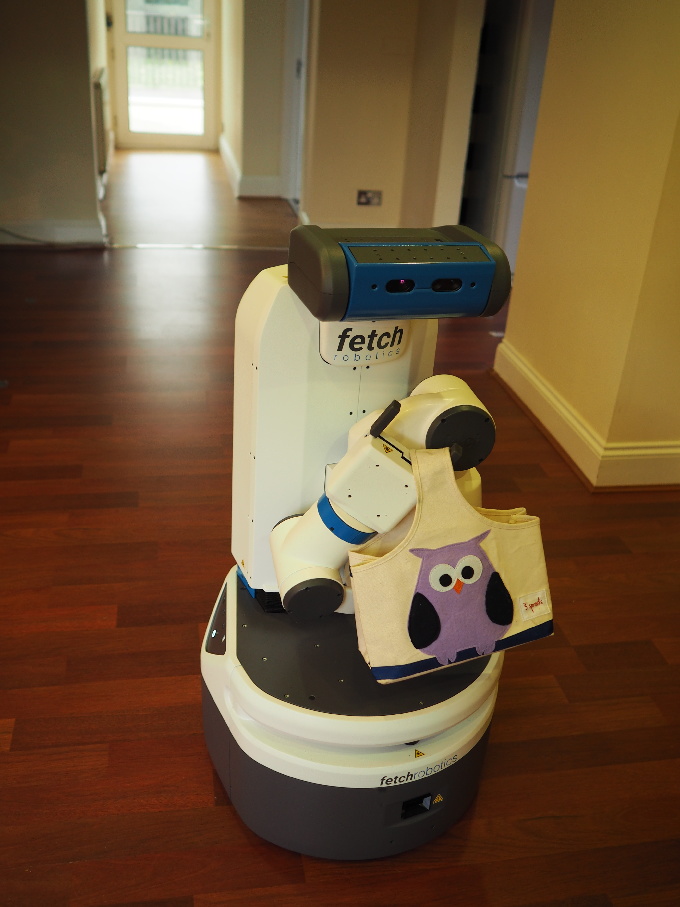}
\caption{The Fetch Robot at Robot House, its arm configuration, and the basket used for the experiment.}\label{fig:fetch-robot}
\end{figure}

The home environment was the University of Hertfordshire's Robot
House, a house people can visit and experience robots and sensors in a
home environment.  We used a Fetch Robot (hereafter Fetch)
(Fig.~\ref{fig:fetch-robot}). Fetch was selected as it can move at a
rate of 1 meter per second (m/s). This speed is slower than an average
person’s walking speed of 1.4~m/s
\parencite{RoseEnergeticsWalking2006}, but accelerating up to this
speed takes enough time that it is possible to create different
velocity profiles.  Fetch held a basket to reduce uncertainty in the
handover.

Fetch used its own navigation software to navigate to the different
points, which is based on the navigation stack from the Robot
Operating System (\abbr{ROS})
\parencite{OpenSourceRoboticsFoundationROS2019}. We chose to use the
navigation stack as we trusted the stack had better safeguards to move
the robot around and avoid obstacles than blind velocity commands.
There were pre-assigned destinations to move to in the house:
(Positions~1, 2, and 3 respectively on Fig.~\ref{fig:floor-plan}).
Each spot had two poses, one for facing the person and one for facing
away from the person towards the next location. The two poses per
location was done to keep the performance of Fetch's navigation
software similar across conditions. Position~1 had a slightly
different locations for its poses to make it easier to remove and add
items to the basket without the participant noticing.

In an effort to improve the internal validity of the experiment, we
did not want confounding factors such as speech recognition or sound
to be part of the experiment. We only wanted to use motion. When
Fetch had arrived at the pose facing the person, it would raise its
torso 10 cm.\ to indicate that it was ready to receive a cup, or that
the person should take (and later return) the questionnaire. This
movement—one could argue a form of configuration legibility—was added
after a pilot study showed that a participant could be confused as to
when the participant should give the cup to  the robot.

For the velocity profiles, we adapted the algorithm from
\textcite{SchulzClassifyingHumanRobot2018}
to be a plugin for the local planner in the
 navigation system. The
plugin was a modified version of Fetch's local planner and was based on
the trajectory roll out scheme
\parencite{GerkeyPlanningcontrolunstructured2008}. This was similar to
a suggested set up for integrating stylized motion into a velocity
profile for a task
\parencite{GielniakStylizedmotiongeneralization2010}. The plugin
included an additional parameter that can be modified dynamically for
setting the velocity profile (linear or \siso). This allowed us to
quickly change the velocity profile without having to restart the
robot's navigation system. The changes only affected Fetch's linear
velocity (i.e., moving forward); the angular velocity (i.e., turning
in place) was always based on a linear velocity profile.

\begin{figure*}[htb]
\centering
\includegraphics[width=0.49\textwidth]{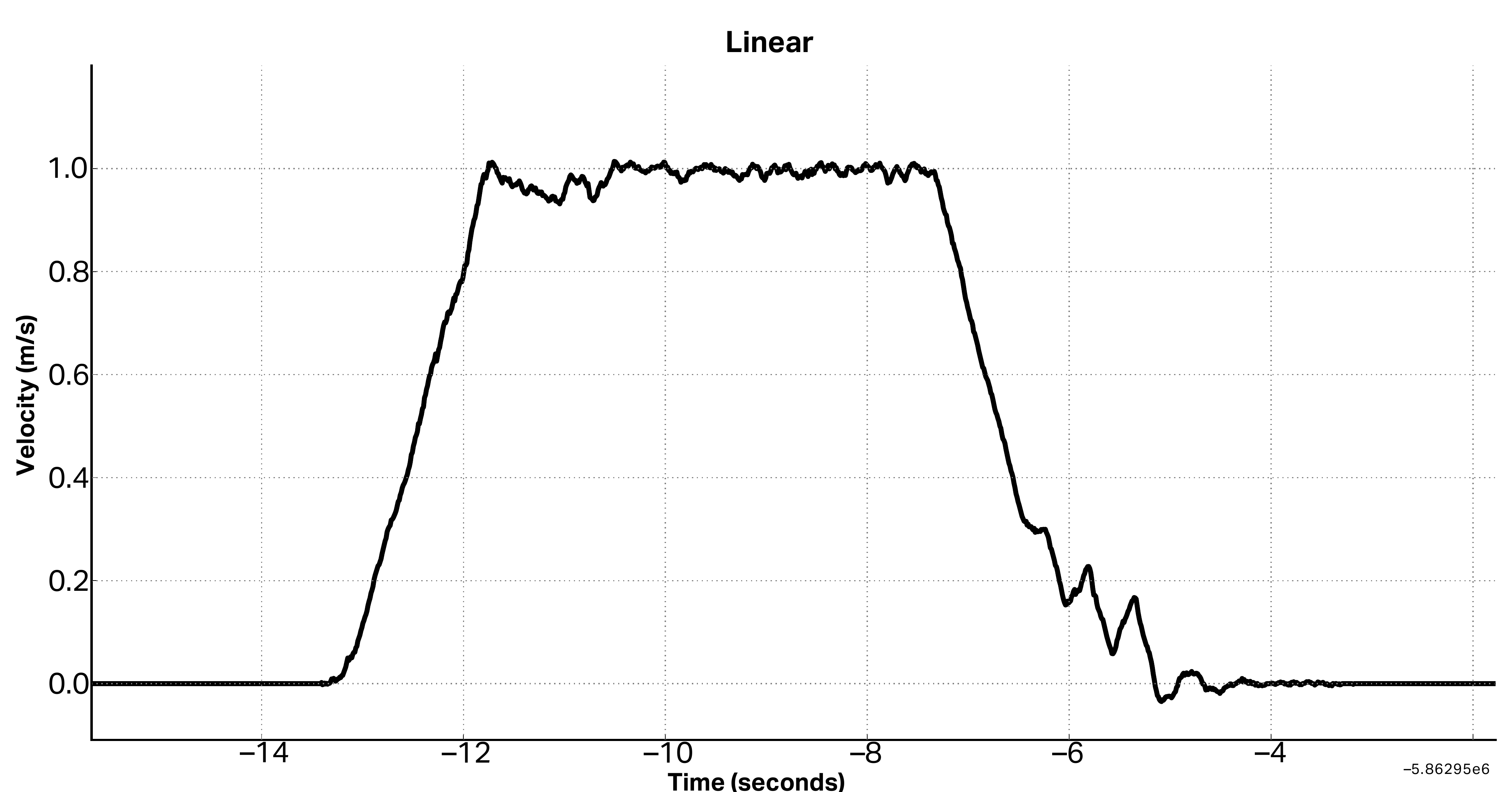}
\includegraphics[width=0.49\textwidth]{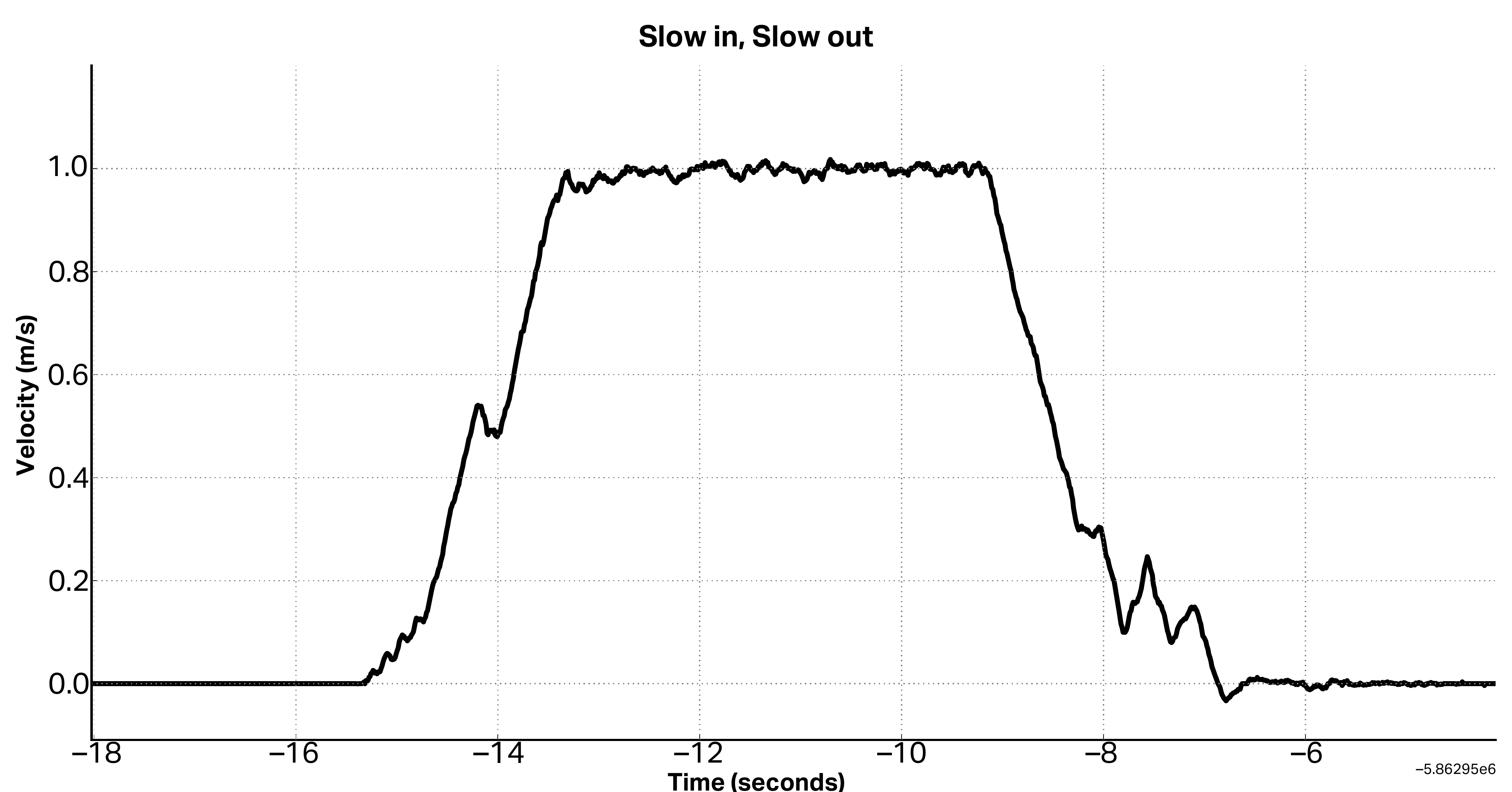}
\caption{Odometry samples of the robot's linear velocity when moving from Position~3 to Position~1 for linear (\textit{left}) and \siso (\textit{right}). Data line has been thickened for readability at this size. The time spent moving from start to finish is similar.}\label{fig:odom-compare}
\end{figure*}

Fetch was \emph{partially} controlled via Wizard of
Oz. The Wizard of Oz component consisted of the facilitator in the
kitchen (the paper's first author) watching when the participant and
the facilitator on the couch had put a cup in Fetch's basket, and then
instructing Fetch to navigate to the next Position. This allowed us to
reduce time variability of the experiment; we also did not have time to
implement adding sensors and code for checking that the cups had been put
in the basket. The choice of having a person in the kitchen and the
couch also allowed two people to watch the robot and activate an
emergency stop if Fetch was going to run into something.

We collected additional data from Fetch. This included the robot's
odometry information (e.g., Fig.~\ref{fig:odom-compare}) and the time
from when a request to move was made to move to the next location
until the time that the robot arrived at the location and raised its
torso.

\section{Results}\label{sec:results}

There were 38 participants who performed the experiment. 19 of the
participants identified as female and 19 as male. Ages range from
18 to 80 years (mean age:~37.39 years, median age:~34.5 years,
\abbr{SD}:~15.74 years). 16 participants had no experience with
robots. Each participant had two interactions with the linear velocity
profile and two interactions with the \siso profile for
a total of 152 encounters (76 linear and 76 \siso).

We collected quantitative and qualitative data, but due to space
constraints, we will only present the Godspeed series results.  Since the
Godspeed Perceived Safety has two items where the negative item is at
the top of the scale and the positive item at the bottom of the scale,
we have reversed those items for the calculations presented below.

\subsection{Tests for Data Quality}

The main measure was the Godspeed series.  In
introducing the series, practitioners were advised
to use Cronbach’s $\alpha$ to test for the internal consistency of the
questions and to see if the responses were reliable for each series of
questions
\parencite{BartneckMeasurementInstrumentsAnthropomorphism2009}. The
results are in Table~\ref{tab:crombach-alpha}.

\begin{table}[htb]
\caption{Cronbach's $\alpha$ for each Godspeed series}\label{tab:crombach-alpha}
\center{}
\begin{tabular}{lS[table-format=1.2]}
\toprule
\textbf{Godspeed series} &
\textbf{Cronbach's $\alpha$} \\
\midrule
Anthropomorphism &
0.86 \\
Animacy &
0.84 \\
Likeability &
0.9 \\
Perceived Intelligence &
0.84 \\
Perceived Safety &
0.63 \\
\bottomrule
\end{tabular}

\end{table}

The rule of thumb is a Cronbach's $\alpha$ over 0.7 has good
internal consistency. The Anthropomorphism, Animacy, Likeability, and
Perceived Intelligence series exceed this threshold and should
have good consistency (Table~\ref{tab:crombach-alpha}). The Perceived
Safety series is below the 0.7 threshold at 0.63. This series
has fewer items than the other series (three versus five or more for
the others). However, the score indicates a fair
agreement among the perceived safety variables. So, the Godspeed
Perceived Safety was still considered below.

We then ran a Shapiro-Wilk normality test to see if the responses were
a normal distribution. The test was run on all the Godspeed semantic
items, and results indicated the data was \emph{not} a normal
distribution (all \(p<.0000001\)). However, running the test on the
average of items per series indicated that the average for
Anthropomorphism series may be a normal distribution
(\(p>.05\)). Regardless, the Shapiro-Wilk test indicated that we
should employ statistical tests that did not assume a normal
distribution like the Wilcoxon Matched-Pairs Signed-Ranks test when
checking statistical significance, and bootstrapping when calculating
confidence intervals (\abbr{CI}).

\subsection{Hypothesis Testing}

\begin{figure}[htb]
\centering
\includegraphics[width=0.50\textwidth]{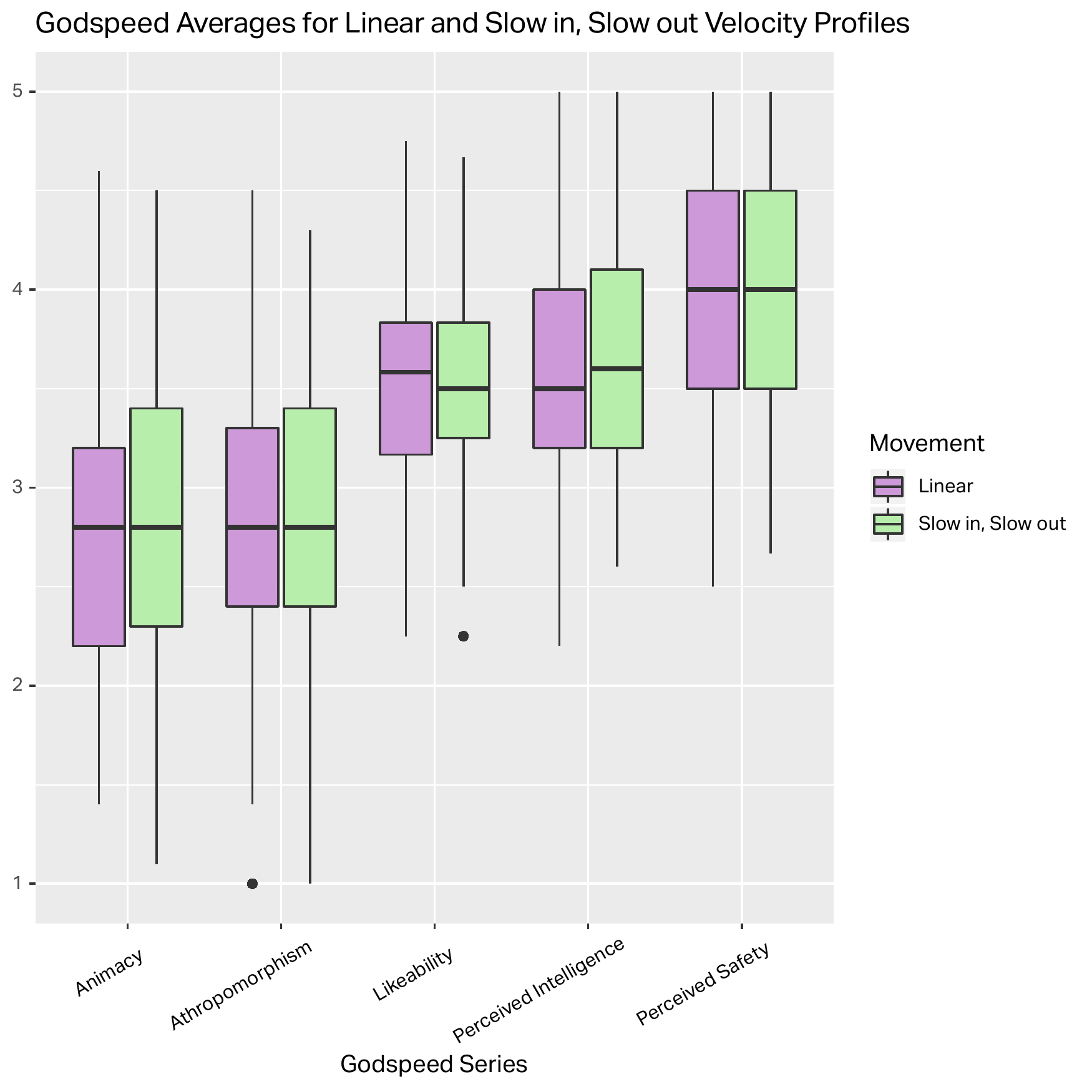}
\caption{Box plot of the averages for each Godspeed Series, split by \SiSo and Linear.}\label{fig:gs-avg-box}
\end{figure}

\begin{table}[htb]
  \caption{Number of participants with complete responses, actual and bootstrapped mean, and Bootstrap Bias-Corrected Accelerated 95\% Confidence interval for Godspeed series used in experiment, split by Linear and \SiSo velocity profile.}\label{tab:gs-avg}
  \center{}
  \begin{tabular}{lrS[table-format=1.2]S[table-format=1.2]c}
    \toprule
    \textbf{Godspeed series} & $n$ & \multicolumn{1}{c}{\textbf{Mean}} & \multicolumn{1}{c}{\textbf{Bootstrap}} & \textbf{95\% \abbr{BC}a \abbr{CI}} \\
    \midrule
    \textbf{Anthropomorphism} \\
    Linear & 36 & 2.84 & 2.84 & (2.56, 3.06) \\
    \SiSo & 37 & 2.83 & 2.83 & (2.60, 3.08) \\
    \midrule
    \textbf{Animacy} \\
    Linear & 37 & 2.88 & 2.88 & (2.65, 3.1) \\
    \SiSo & 38 & 2.87 & 2.87 & (2.66, 3.1) \\
    \midrule
    \textbf{Likeability} \\
    Linear & 37 & 3.59 & 3.59 & (3.37, 3.82) \\
    \SiSo & 38 & 3.71 & 3.71 & (3.52, 3.92) \\
    \midrule
    \textbf{Perceived Intelligence} \\
    Linear & 37 & 3.57 & 3.57 & (3.36, 3.77) \\
    \SiSo & 37 & 3.51 & 3.51 & (3.32, 3.69) \\
    \midrule
    \textbf{Perceived Safety} \\
    Linear & 38 & 3.93 & 3.93 & (3.71, 4.12) \\
    \SiSo & 38 & 3.98 & 3.98 & (3.77, 4.19) \\
    \bottomrule
  \end{tabular}
\end{table}

The corresponding null hypotheses for the hypotheses in
Section~\ref{sec:introduction} would be there is no difference between
the perceived qualities of the robot and velocity profiles.
We grouped each participants' encounters by linear and \siso velocity
profile and calculated the average (mean) score for each semantic item
in these groups; missing responses to one or more missing semantic
items were dropped for that series (i.e., \(n=38\) for groups with no
missing items). We then took the mean of all items for each series.
Since this data is not normally distributed, we used bootstrapping
with 2,000 iterations to calculate a 95\% Bias-Corrected accelerated
(\abbr{BC}a) confidence interval on the average. The series'
distribution is presented as a box plot (Fig.~\ref{fig:gs-avg-box}).
The mean scores and \abbr{CI}s are presented in
Table~\ref{tab:gs-avg}.

Table~\ref{tab:gs-avg} shows small differences in the means and large
overlap in the \abbr{CI}s. Fig.~\ref{fig:gs-avg-box} shows differences
in terms of the extremes and the grouping for likeability and
perceived intelligence, but is there a statistically significant
difference? Wilcoxon Matched-Pairs Signed-Ranks tests were run on the
means to see if we could reject the null hypotheses. Since the
hypotheses are comparisons in a family, we ran a Holm-Bonferroni
correction to check that multiple comparisons do not result in
something randomly becoming significant. The Holm-Bonferroni
correction indicated none of the groupings were statistically
significant (all \(p>.05\)). This, combined with the \abbr{CI}s,
indicate we cannot reject the null hypotheses. These results were not
appreciably different when only looking at participants who had no
experience with robots.

\section{Discussion}\label{sec:discussion}

Altering a velocity profile can have an effect on the safety of the
robot. We expected to see a pronounced effect between the velocity
profiles, but this was not the case. However, presenting these results
are important for other \hri researchers to know that a certain
technique does not provide a certain effect. Using animation
techniques with robots are under investigation and used in other
studies as presented in Section~\ref{sec:related-work}, and will
likely continue to do so. Knowing what works or what does not work
helps this research.

This also provides an opportunity to help in the design of
future studies. One place to look for future research directions is to
review responses to the different semantic items of the Godspeed Series. We can
also reexamine the experiment to see if there are other items that
need to be considered.

\subsection{Semantic Item Exploration}

Although we could not reject the null hypotheses, we explored the
semantic items for cases where there might be
differences. As this study was not specifically looking at these
comparisons, we do \emph{not} draw conclusions,
but they may point to possible items to explore in future studies.

\begin{figure}[htb]
\centering
\includegraphics[width=0.50\textwidth]{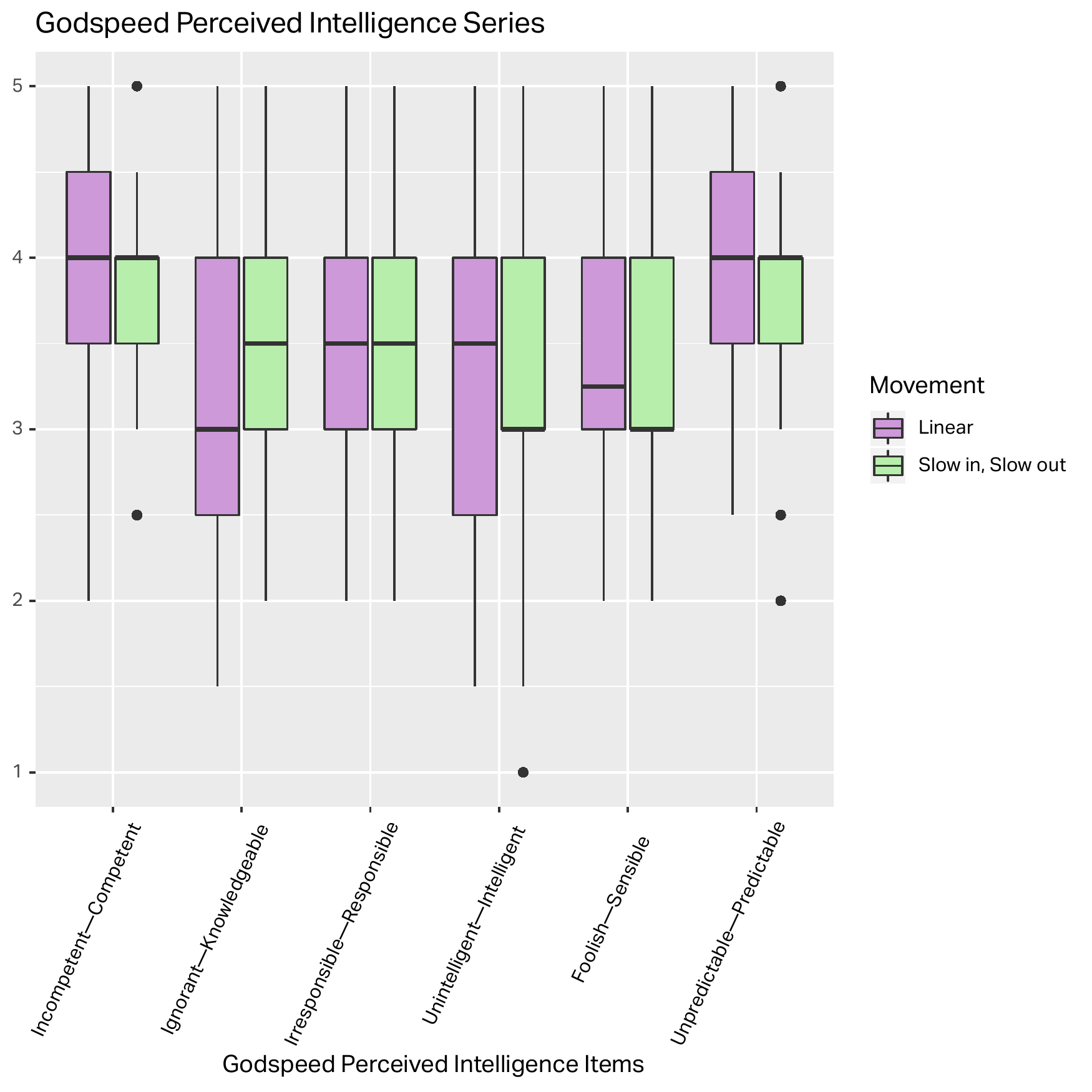}
\caption{Box plot of the averages the Perceived Intelligence semantic items, split by \SiSo and Linear.}\label{fig:gs-intelligence-box}
\end{figure}

\begin{figure}[htb]
\centering
\includegraphics[width=0.50\textwidth]{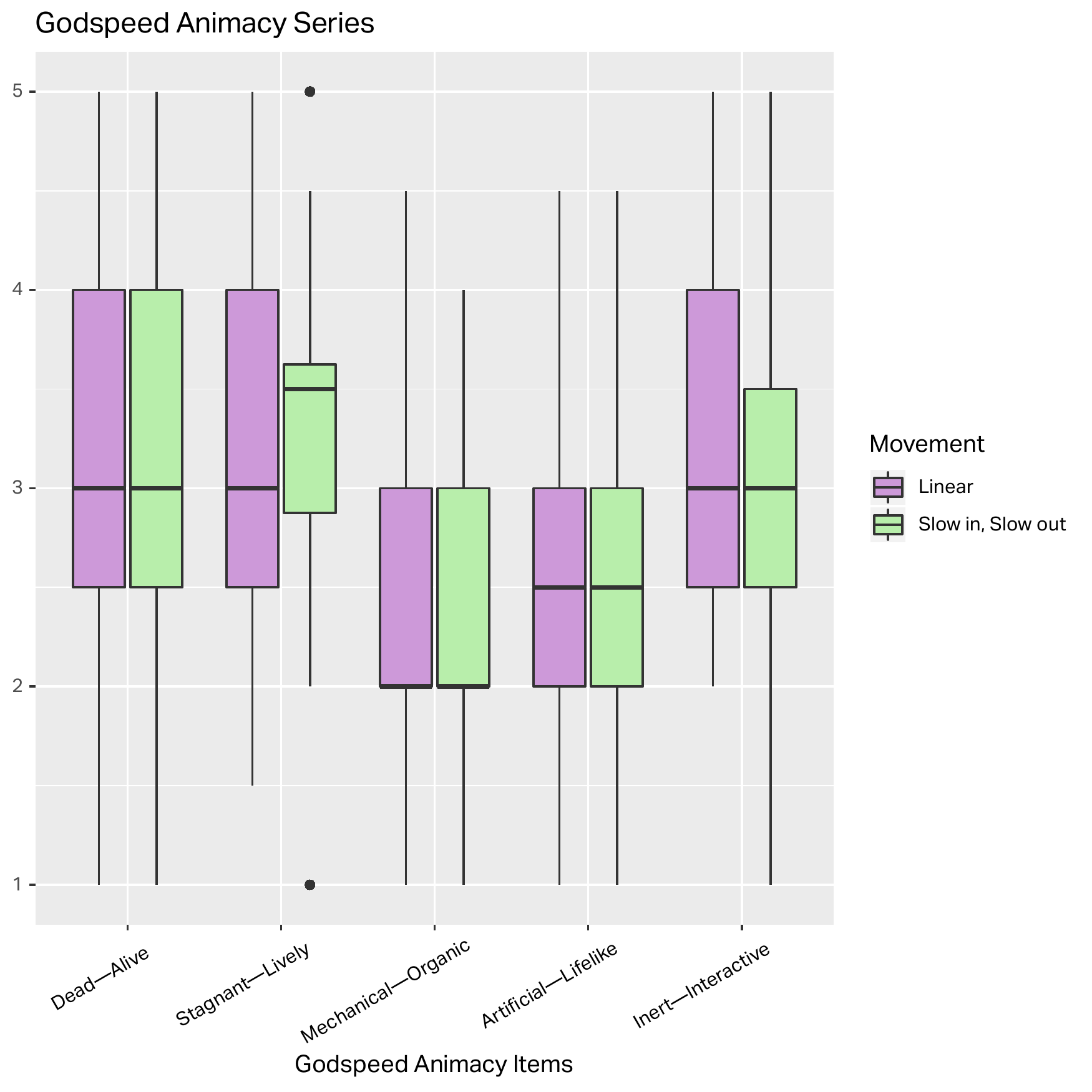}
\caption{Box plot of averages of Animacy semantic items; the participants who received the wrong Inert—Interactive item are removed.}\label{fig:gs-animacy-box}
\end{figure}

\begin{figure}[htb]
\centering
\includegraphics[width=0.50\textwidth]{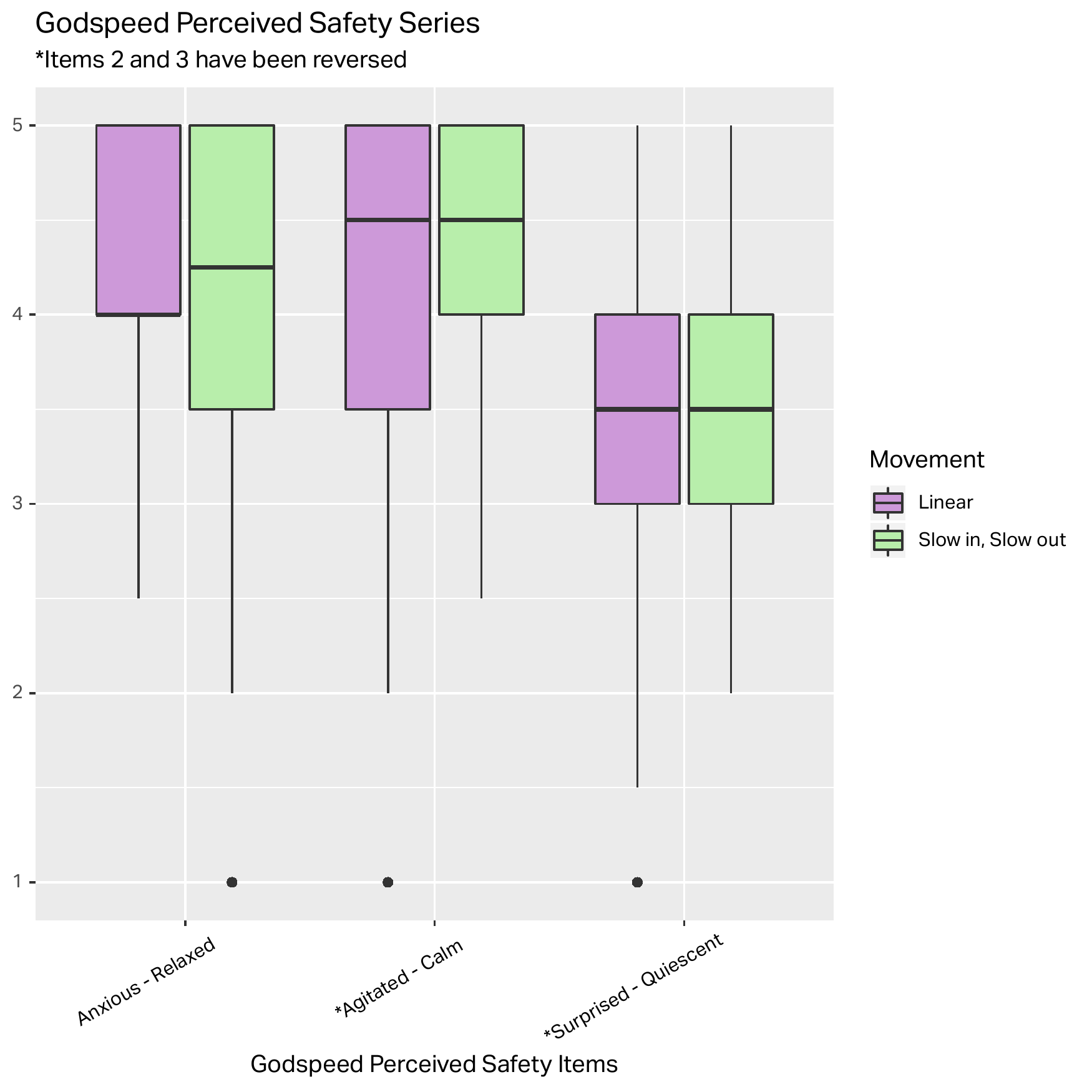}
\caption{Box plot of the averages the Perceived Safety semantic items, split by \SiSo and Linear.}\label{fig:gs-safety-box}
\end{figure}

We ran Wilcoxon Matched-Pairs Signed-Ranks tests and bootstrapped 95\% confidence intervals on the
semantic items. This showed three interesting items: the
Unpredictable—Predictable item for the Perceived Intelligence Series
(Fig.~\ref{fig:gs-intelligence-box}, Table~\ref{tab:gs-intelligence}),
the Inert—Interactive item for the Animacy series
(Fig.~\ref{fig:gs-animacy-box}, Table~\ref{tab:gs-animacy}), and the
Calm—Agitated item for the Perceived Safety series
(Fig.~\ref{fig:gs-safety-box}, Table~\ref{tab:gs-safety}).

\begin{table}[htb]
  \caption{Participants with complete responses, actual and bootstrapped mean, and 95\% \abbr{BC}a \abbr{CI} for Godspeed Perceived Intelligence semantic items used in experiment, split by Linear and \SiSo velocity profile.}\label{tab:gs-intelligence}
  \center{}
  \begin{tabular}{p{2.9cm}rS[table-format=1.2]S[table-format=1.2]c}
    \toprule
    \textbf{Perceived Intelligence Item} & $n$ & \multicolumn{1}{c}{\textbf{Mean}} & \multicolumn{1}{c}{\textbf{Bootstrap}} & \textbf{95\% \abbr{BC}a \abbr{CI}} \\
    \midrule
    \textbf{Incompetent—Competent} \\
    Linear & 37 & 3.84 & 3.84 & (3.55, 4.04) \\
    \SiSo & 37 & 3.78 & 3.78 & (3.58, 3.96) \\
    \midrule
    \textbf{Ignorant—Knowledgeable} \\
    Linear & 38 & 3.28 & 3.28 & (3.00, 3.54) \\
    \SiSo & 38 & 3.33 & 3.32 & (3.09, 3.55) \\
    \midrule
    \textbf{Irresponsible—Responsible} \\
    Linear & 38 & 3.55 & 3.55 & (3.38, 3.88) \\
    \SiSo & 38 & 3.63 & 3.63 & (3.32, 3.79) \\
    \midrule
    \textbf{Unintelligent—Intelligent} \\
    Linear & 38 & 3.39 & 3.40 & (3.09, 3.67) \\
    \SiSo & 38 & 3.30 & 3.31 & (3.00, 3.59) \\
    \midrule
    \textbf{Foolish—Sensible} \\
    Linear & 38 & 3.47 & 3.47 & (3.19, 3.74) \\
    \SiSo & 38 & 3.38 & 3.38 & (3.13, 3.61) \\
    \midrule
    \textbf{Unpredictable—Predictable} \\
    Linear & 38 & 3.96 & 3.96 & (3.71, 4.16) \\
    \SiSo & 38 & 3.70 & 3.70 & (3.42, 3.89) \\
    \bottomrule
  \end{tabular}
\end{table}

\begin{table}[htb]
  \caption{Participants with complete responses, actual and bootstrapped mean, and 95\% \abbr{BC}a \abbr{CI} for Godspeed Animacy semantic items used in experiment, split by Linear and \SiSo velocity profile; the participants who received the wrong Inert—Interactive item are removed.}\label{tab:gs-animacy}
  \center{}
  \begin{tabular}{lrS[table-format=1.2]S[table-format=1.2]c}
    \toprule
    \textbf{Animacy Item} & $n$ & \multicolumn{1}{c}{\textbf{Mean}} & \multicolumn{1}{c}{\textbf{Bootstrap}} & \textbf{95\% \abbr{BC}a \abbr{CI}} \\
    \midrule
    \textbf{Dead—Alive} \\
    Linear & 31 & 3.18 & 3.18 & (2.77, 3.53) \\
    \SiSo & 32 & 3.19 & 3.19 & (2.83, 3.50) \\
    \midrule
    \textbf{Stagnant—Lively} \\
    Linear & 31 & 3.27 & 3.28 & (2.95, 3.56) \\
    \SiSo & 32 & 3.23 & 3.24 & (2.91, 3.50) \\
    \midrule
    \textbf{Mechanical—Organic} \\
    Linear & 31 & 2.34 & 2.34 & (2.15, 3.66) \\
    \SiSo & 32 & 2.41 & 2.41 & (2.15, 2.65) \\
    \midrule
    \textbf{Artificial—Lifelike} \\
    Linear & 31 & 2.52 & 2.51 & (2.21, 2.79) \\
    \SiSo & 32 & 2.52 & 2.52 & (2.23, 2.81) \\
    \midrule
    \textbf{Inert—Interactive} \\
    Linear & 31 & 3.13 & 3.13 & (2.87, 3.39) \\
    \SiSo & 32 & 3.03 & 3.03 & (2.67, 3.38) \\
    \bottomrule
  \end{tabular}
\end{table}

\begin{table}[htb]
  \caption{Participants with complete responses, actual and bootstrapped mean, and 95\% \abbr{BC}a \abbr{CI} for Godspeed Perceived Safety semantic items used in experiment, split by Linear and \SiSo velocity profile.}\label{tab:gs-safety}
  \center{}
  \begin{tabular}{lrS[table-format=1.2]S[table-format=1.2]c}
    \toprule
    \textbf{Perceived Safety Item} & $n$ & \multicolumn{1}{c}{\textbf{Mean}} & \multicolumn{1}{c}{\textbf{Bootstrap}} & \textbf{95\% \abbr{BC}a \abbr{CI}} \\
    \midrule
    \textbf{Anxious—Relaxed} \\
    Linear & 38 & 4.20 & 4.20 & (3.93, 4.42) \\
    \SiSo & 38 & 4.07 & 4.06 & (3.69, 4.33) \\
    \midrule
    \textbf{Agitated—Calm} \\
    Linear & 38 & 4.12 & 4.11 & (3.74, 4.38) \\
    \SiSo & 38 & 4.30 & 4.30 & (4.03, 4.51) \\
    \midrule
    \textbf{Surprised—Quiescent} \\
    Linear & 38 & 3.47 & 3.47 & (3.18, 3.71) \\
    \SiSo & 38 & 3.57 & 3.56 & (3.30, 3.80) \\
    \bottomrule
  \end{tabular}
\end{table}

Looking first at the Unpredictable—Predictable series, we suspect
there was a confounding factor that caused the difference in being
unpredictable. When preparing the experiment, we found no noticeable
difference in the time for the robot to move between
positions. Fig.~\ref{fig:odom-compare} shows the times for Fetch to
travel from Position~3 to Position~1 are within a half second between
velocity profiles.

However, during the experiment the averages of the timings we recorded
shows a large difference between instructing the robot to move from
Position~3 to Position~1 and the robot completing the move. The time
the robot used in preparing for this move sometimes took much longer
(on average about five seconds longer for the \siso velocity profile).
In these situations, Fetch's navigation system would register Fetch as
``stuck'' and rotate the robot to help reset its local obstacle map.
This would add more time from when the robot was requested to move
until it completed its move. This happened more often for the \siso
condition than the linear condition (17 out of 76 times versus 4 out
of 76 times) and may explain the difference. Many participants noted
this rotating when filling out the questionnaires and this could also
have resulted in the perception in predictability. We are unsure what
caused this ``stuck'' behavior. We suspect that it was latency with
the wireless network combined with delays in the navigation system
being run on a different computer and not Fetch. This did not turn up
in our pilot, but we should have accounted for the possibility of this
behavior.

Examining the Intert—Interactive semantic item for the Animacy series
had problems as its semantic scale was incorrect for 6 participants
(Due to a typo, the scale was Inert—\emph{Inactive} instead of
Inert—\emph{Interactive}). This was corrected for the other 32
participants. When these 6 participants were removed, the resulting
test shows no statistically significant difference.

The box plot for the Perceived Safety series
(Fig.~\ref{fig:gs-safety-box}) shows more concentration at the calm
end of the semantic item for the \siso versus the linear velocity
profile, and Table~\ref{tab:gs-safety} indicates the average may be
different. However, the Wilcoxon Signed Ranks test is a test of how
the spread of the scores match, \emph{not} the match in average score.
Figs.~\ref{fig:gs-intelligence-box}–\ref{fig:gs-safety-box} show that
the median scores of Distracted—Calm, Unpredictable—Predictable, and
Inert—Interactive semantic items are the same between the Linear and
\SiSo velocity profiles. Future work would need to examine whether the
concentration is real and desirable. Other items did not show a
statistically significant difference and had similar averages, but had
different medians (e.g., Stagnant—Lively of the Animacy Series in
Fig.\ref{fig:gs-animacy-box} and Table~\ref{tab:gs-animacy}).

\subsection{Examining the Scenario}

The scenario we described can be used by others also
wishing to examine aspects of a robot—like movement—while the person
is engaged in a task with the robot in a home context. Participants in
the experiment generally watched the robot most of the time it was
moving, so this might be useful for studies requiring participants to watch a robot.

Most animation work consists of several principles. One reason we ran
this study was to see how powerful one principle of animation on its
own could be in a scenario. That is, if we only employ one principle
for a robot in a real world scenario, does it make a difference in
people's perceptions? In this scenario, the \siso velocity profile is
not enough to change perceptions of animacy, anthropomorphism,
likeability, perceived intelligence, or perceived safety on its own.
But there might be other factors that are overpowering the
principle. For example, the movement of the robot's torso and robot's
turning did \emph{not} use \siso. This may mean that these other
velocity profiles overpowered the robot's movement from the different
positions. In addition, Fetch's momentum also created a type of \siso
effect even in the linear velocity profile (visible in
Fig.~\ref{fig:odom-compare}). Events such as the robot getting
``stuck'' may also have a larger impact than the \siso velocity
profile on its own. Even other properties of Fetch—such as appearance,
color, or texture appearance may overpower how it moves. So this
raises a question about how much the result is tied to the specific
robot for the scenario.

As mentioned in Section~\ref{sec:method}, people will likely be
interacting with a robot, not simply watching how it moves. The \siso
velocity profile may be too subtle on its own for people to easily
perceive the difference while working on the task. Perhaps changes can
be made to the velocity curves to make them more distinct, fixing the
``stuck'' events, a different approach angle, or a different activity
could make the differences easier to notice.

It may also be that animation principles must be combined to get a
noticeable effect. Participants in a study of a greeting robot
that used a straight movement versus movement based on the animation
principle of \principle{arcs} did not mention any difference between
these two types of movements in their interviews
\parencite{Anderson-BashanGreetingMachineAbstract2018}. Further
research is needed to see how combining principles can lead to
noticeable effects and if the noticeable effects are desired in a
robot's design.

Another issue can be that Godspeed questionnaire may not be the best
tool to capture the perceptions here. Our exploration of the semantic
items suggest examining safety, legibility, and predictability with
the \siso velocity profile. A different measurement tool, perhaps
based on the perceived safety scale, may provide a better answer in a
future version of this scenario. On the other hand, others have
suggested that a subjective measurement scale may not be sufficient
for capturing people's perceptions, and the results could differ from
observed data
\parencite{WinkleEffectivePersuasionStrategies2019}. Qualitative data
may also provide a richer picture in these circumstances. We are examining
the qualitative data from our participants and will present this in
the future. Different tools and a study focusing on one element may
provide stronger statistical power on the quantitative side, and a
deeper understanding on the phenomena on the qualitative side.

This set up also shows the need to carefully decide hypotheses
for quantitative measures. Adding a measure (here,
an additional Godspeed series) broadens the scope of what you can
find, but it adds additional requirements on rejecting a null
hypothesis. It is important to consider
using descriptive statistics to help explain the
effect the data is showing \parencite{Baxtercharacterisingthreeyears2016}.

\section{Conclusion and Future Work}

We ran a within-subjects survey looking at how a person perceives a robot
moving with a linear velocity profile
versus a profile based on the \siso animation principle, while
performing a task together.  We used the Godspeed Questionnaire series to
measure the perceptions of anthropomorphism, animacy, likeability,
perceived intelligence, and perceived safety. The average scores of
these series indicate that there is not enough difference to indicate
an effect.
We have documented our study and provided the
implementation for the velocity curves, the software for running the
experiment, and for analyzing
results\footnote{https://gitlab.com/robothouse/rh-experiments/siso}. We
are examining the qualitative data to see if we can create
a more complete picture of people's perceptions.

Even without any major perceived distinction between the two profiles,
there are still possibilities to explore with the \siso velocity
profile. Examining the Godspeed semantic items shows that aspects of
participants' perceived safety may be influenced with a \siso velocity
profile. Future work could examine if a \siso velocity profile
provides a better indication of a robot starting and stopping and thus
makes the robot's locomotion more legible or predictable to the humans
interacting with it. Another aspect would be combining \siso with
other principles like \principle{anticipation} to further enhance this
legibility. A focused study on these topics, supplemented with other
research methods, can provide insight into how humans and robots can
work together safely.

\section*{Acknowledgment}

We thank Robot House at the University of Hertfordshire
for allowing us to use their facilities to design and run the
experiment described in this paper. We also thank Ragnar Hauge for
suggestions and help on statistical methods.


\printbibliography[]

\end{document}